\documentclass[conference]{IEEEtran}
\IEEEoverridecommandlockouts
\usepackage{amsmath,amssymb,amsfonts}
\usepackage{makecell}
\usepackage{multirow}
\usepackage[skip=0pt, font={footnotesize}, labelsep=period]{caption}
\usepackage{subcaption}
\usepackage{algorithmic}
\usepackage{graphicx}
\usepackage{textcomp}
\usepackage{xcolor}
\def\BibTeX{{\rm B\kern-.05em{\sc i\kern-.025em b}\kern-.08em
    T\kern-.1667em\lower.7ex\hbox{E}\kern-.125emX}}
\renewcommand{\baselinestretch}{0.98}
\begin{document}

\title{Design Methodology for Deep Out-of-Distribution Detectors in Real-Time Cyber-Physical Systems}
% Design framework for Deployment of VAE-based OOD Detectors in Real-Time Systems

\author{\IEEEauthorblockN{Michael Yuhas, Daniel Jun Xian Ng, Arvind Easwaran}
\IEEEauthorblockA{\textit{Nanyang Technological University,}
Singapore \\
michaelj004@e.ntu.edu.sg, danielngjj@ntu.edu.sg, arvinde@ntu.edu.sg}
}

\maketitle

\begin{abstract}
When machine learning (ML) models are supplied with data outside their training distribution, they are more likely to make inaccurate predictions; in a cyber-physical system (CPS), this could lead to catastrophic system failure.  To mitigate this risk, an out-of-distribution (OOD) detector can run in parallel with an ML model and flag inputs that could lead to undesirable outcomes.  Although OOD detectors have been well studied in terms of accuracy, there has been less focus on deployment to resource constrained CPSs. In this study, a design methodology is proposed to tune deep OOD detectors to meet the accuracy and response time requirements of embedded applications.  The methodology uses genetic algorithms to optimize the detector's preprocessing pipeline and selects a quantization method that balances robustness and response time. It also identifies several candidate task graphs under the Robot Operating System (ROS) for deployment of the selected design. The methodology is demonstrated on two variational autoencoder based OOD detectors from the literature on two embedded platforms. Insights into the trade-offs that occur during the design process are provided, and it is shown that this design methodology can lead to a drastic reduction in response time in relation to an unoptimized OOD detector while maintaining comparable accuracy.
\end{abstract}
\vspace{-2mm}

%\begin{IEEEkeywords}
%Cyber-physical systems, out-of-distribution data, response time optimization, deep neural networks, genetic algorithms.
%\end{IEEEkeywords}

% Introduction (1 pages)
% What is an OOD? What is the motivation / problem? The need for a design exploration framework... 
% Missing a figure to motivate the problem. 
\section{Introduction}
\vspace{-1mm}
When machine learning (ML) algorithms are presented with inputs that lie outside of their training distribution, they are unlikely to make accurate predictions.  In a safety-critical cyber-physical system (CPS) such as an autonomous vehicle (AV), inaccurate predictions can lead to severe consequences.  After the ML model is trained and deployed, there is no guarantee that the samples encountered in the field lie within the distribution of training data.  If these out-of-distribution (OOD) samples can be detected, control can be returned to a human operator or the system can enter a safe state. Consider a YOLO object detection network deployed to an AV.  If the vehicle is being operated in conditions that do no reflect its training data (e.g., extreme rain), it is unlikely that the YOLO network will be able to accurately identify obstacles. The OOD detector deployed to this vehicle needs to accurately identify when OOD samples are present and perform this detection within a maximum response time.  For example, if the vehicle is traveling at 100~kph and OOD detection takes 1~s, the vehicle will have traveled 28~m before it can react.  However, if OOD detection takes 100~ms, the vehicle will have only travelled 2.8~m and there will be additional time to plan an alternate route, notify the human driver, or decelerate.

\begin{figure}
    \centering
    \includegraphics[width=0.475\textwidth]{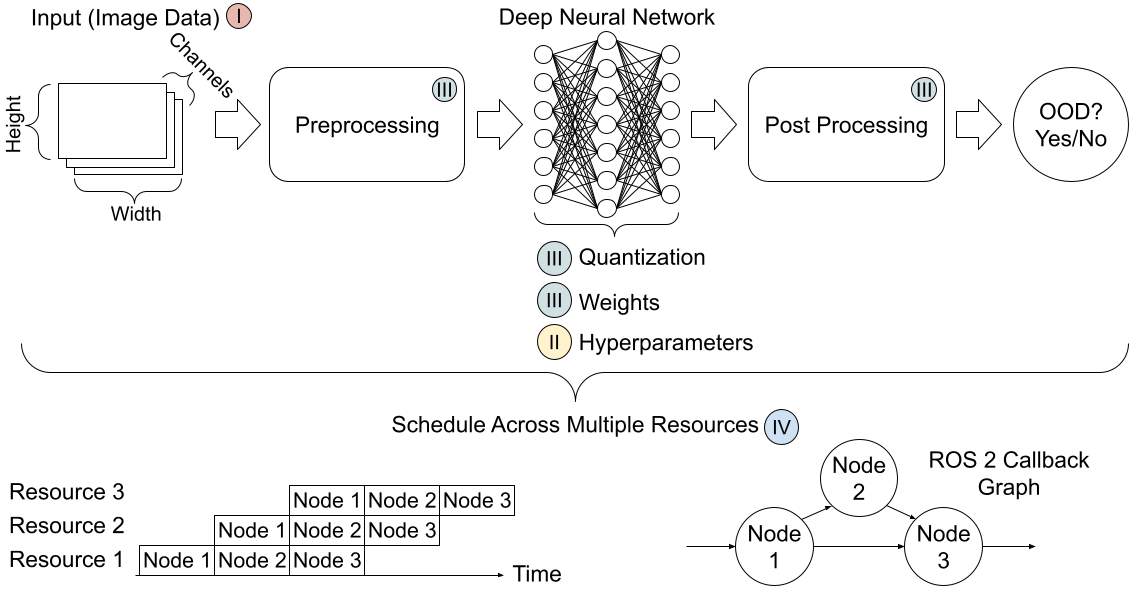}
    \caption{Components of a generic deep OOD detector.  Image input data is preprocessed and supplied to a DNN, the output undergoes post processing to classify the image as OOD or ID.  This structure can be split into arbitrary ROS callback graphs and scheduled across computational resources.}
    \label{fig:generic}
    \vspace{-7mm}
\end{figure}

Accurate detection of OOD samples has been the focus of previous research, and in this work we consider two such representative OOD detectors. In \cite{ramakrishna2021efficient} a $\beta$-variational autoencoder ($\beta$-VAE) based OOD detection architecture is proposed that uses a deep generative model to identify samples as OOD based on one or more generative factors such as light intensity, rain, background, etc. In \cite{feng2021improving} a deep OOD detector based on optical flow (OF) is proposed to detect OOD samples caused by observed motion in time-series image data.  In general, deep OOD detectors require three major components as shown in Fig.~\ref{fig:generic}: preprocessing, deep neural network inference, and post processing to derive an OOD score.  Each component has many tunable parameters, and changing one can affect the end-to-end performance of the OOD detector. Additionally, the response time of the OOD detection task determines how fast the system can react to unexpected conditions; thus, even a robust OOD detector cannot be deployed to a CPS if its response time exceeds the system's timing constraints \cite{yuhas2021embedded}.  Middleware such as the Robot Operating System (ROS) can be used to map tasks in an OOD detector to an execution graph.  This mapping, along with the parameters from the three stages of OOD detection affect the response time of an OOD detector.

To navigate this complex design space, we propose a design methodology for OOD detectors in CPS that:
%\vspace{-2mm}
\begin{enumerate}
    \item Provides a systematic method to tune an OOD detector to satisfy functional requirements.
    \item Provides guidance on satisfying nonfunctional requirements by exploring the trade-offs between accuracy and response time under various quantization schemes.
    \item Provides candidate ROS callback graphs with different execution characteristics for easy integration with existing cyber-physical applications.
\end{enumerate}
We apply the methodology on two case studies and our experiments show a $37.5\%$ response time reduction with $3.5\%$ AUROC loss for the $\beta$-VAE detector and a $51.2\%$ response time reduction with $0.7\%$ AUROC loss for the OF detector.

\section{Related Work}
\vspace{-1mm}
Previous literature has focused on the deployment of deep neural networks (DNNs) to edge devices. In \cite{hadidi2019characterizing}, DNNs are constructed in popular deep learning frameworks and then deployed to edge devices to compare response time and power consumption.  However, modifications to the image preprocessing pipeline and the trade-off between accuracy and response time were not explored.  In \cite{zhang2019openei}, a framework to aid system designers in selecting a model for edge deployment is proposed.  It accounts for model accuracy, latency, memory footprint, and power usage, but does not propose any tunable parameters to optimize existing models.  In \cite{stacker2021deployment}, both preprocessing of input images and quantization techniques are considered when deploying deep object detection networks to edge devices.  Response time, accuracy, and power usage are compared. In \cite{thar2019meta}, a framework that uses meta learning to select an optimal model for edge deployment is proposed, however, this framework relies on tuning the hyperparameters of a DNN, whereas OOD detectors contain additional preprocessing and post processing stages that also affect accuracy and response time.  A survey of techniques for quantization, hashing, and pruning of general DNNs was conducted in~\cite{berthelier2021deep}.
%\vspace{-1mm}

% Framework (2 pages)
% Functional requirements (AUROC / Accuracy)
% nonfunctional requirements (Power, space, RT etc.) Map this into the phases figure. For e.g., phase 3 should link back to requirements (bring it back to requirements)
\section{OOD Detector Design Methodology for CPS}
\label{framework_big}
%\vspace{-1mm}
%\subsection{Design Requirements for OOD Detectors in CPS}
%\label{challenges}
%\vspace{-1mm}
%Functional requirements can be assessed with a variety of metrics like accuracy, F-score, or AUROC.  In this paper, AUROC will be used as a metric to judge the functional performance of an OOD detector because it provides a good characterization of sensitivity vs. specificity across many threshold values. Nonfunctional requirements considered by our framework are response time and throughput.  As demonstrated in the introduction, response time is guarantees the system's ability to take corrective action and avoid a collision.  Maintaining a high throughput is also desirable as it allows the detector to process an incoming data stream at a higher rate and provide more predictions to downstream components even while latency remains fixed.  Designing for functional requirements alone may adversely affect nonfunctional requirements.  For example, using higher resolution images, or deeper neural networks may lead to an increase in a detector's capacity identify OOD samples, but at the expense of more operations yielding a greater response time.
%\vspace{-1mm}
%\subsection{Proposed Design Framework}
%\label{framework}

\begin{figure*}
    \centering
    \includegraphics[width=1\textwidth]{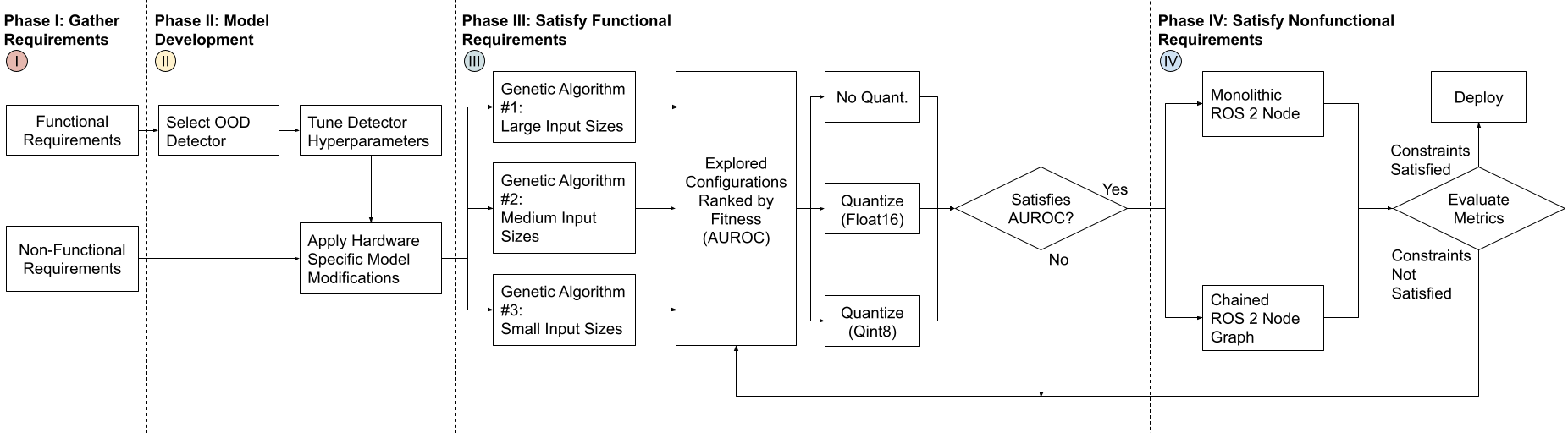}
    \caption{Proposed methodology for deep OOD detector design in CPS.  Colored circles map components from Fig.~\ref{fig:generic} to one or more design phases.}
    \label{fig:framework}
    \vspace{-6mm}
\end{figure*}

Fig.~\ref{fig:framework} shows our proposed OOD detector design methodology.  An exhaustive search across all design parameters would be too costly, so the methodology seeks to generate candidate solutions that satisfy functional requirements and have the potential to satisfy nonfunctional requirements.  The candidate solutions are tweaked until they satisfy all design constraints.

\subsubsection{Gathering Requirements}
Functional requirements for OOD detection are provided in terms of AUROC while nonfunctional requirements such as response time and throughput are also provided.  AUROC is chosen as a metric because of its ability to convey information about the true and false positive rates across an array of thresholds \cite{hendrycks2016baseline}.

\subsubsection{Model Development}
An OOD detector is selected and developed based on an existing architecture. For example, if the goal is to detect OOD samples caused by moving objects in the environment, an OF detector \cite{feng2021improving} could be selected, whereas if the goal is to detect OOD samples caused by specific generative factors, a $\beta$-VAE model \cite{ramakrishna2021efficient} may be more appropriate.  Hyperparameters of the deep network are tuned based on the selected architecture. For an OF detector this means selecting the window size for the Farneback OF algorithm and in a $\beta$-VAE model it means selecting $\beta$ and the number of latent dimensions in the autoencoder.  A list of candidate hardware platforms is also generated based on nonfunctional requirements.  For example, a nonfunctional requirement for 15 fps throughput limits the choice of processors available to designers.  For some hardware in the candidate list, further actions may be needed to adapt the model for deployment.  For example, in this paper's experiments, the Coral TPU \cite{cass2019taking} cannot apply a leaky rectified linear unit (ReLU) activation function to a layer, but can apply a standard ReLU.  In this case, the OOD detectors under evaluation need to be modified to meet the constraints of the underlying hardware platform.

\subsubsection{Satisfying Functional Requirements}
First, an OOD detector's preprocessing pipeline is tuned to generate candidate solutions and afterward quantization is applied.  The goal is to find a set of preprocessing parameters that can reduce network inference time while minimizing the impact on the original model's robustness.  It would not be feasible to explore all parameter combinations, so genetic algorithms (GAs) are used to select the pipeline that leads to the most robust detector. The tunable parameters (i.e., genes) are dependent on the preprocessing steps of the OOD detector. In our experiments population size was selected as 5 and the mutation rate was set at 0.2 to help ensure a good exploration of the parameter space.  These parameters can be adjusted to balance exploring new solutions with the need for quick convergence.  The fitness function was chosen to be the harmonic mean of AUROC for all considered OOD factors to prevent solutions with excellent performance on only one type of OOD.  The search space is divided into small, medium, and large buckets by input image size and each is searched by a separate GA to ensure that solutions likely to lead to a reduction in response time are found. The candidate solutions from each of the GAs are ranked in order of fitness and quantized.  Quantization decreases execution time, but affects the detector's AUROC.  In our experiments, all OOD detectors were trained and tested in Pytorch with 32-bit floating point precision (f32). For our evaluation, we consider three additional quantization levels: half precision (f16), 8-bit static quantization (qint8), and another 8-bit static format specifically for the Coral TPU (Coral) \cite{coraltpuops}.  The quantized models were converted from the f32 models via the model compression process in \cite{feng2021improving}.

\subsubsection{Satisfying Nonfunctional Requirements}

\begin{figure}[htbp]
    \centerline{\includegraphics[width=0.45\textwidth]{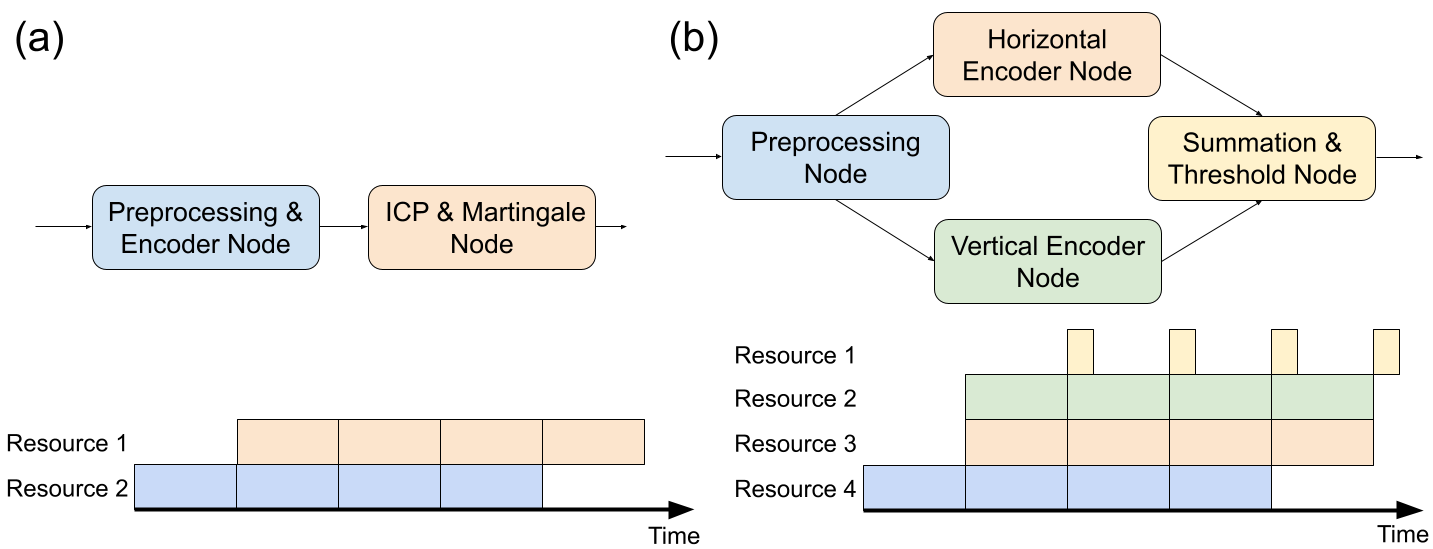}}
    \caption{Callback graphs and ideal resource allocation for the chained OOD detector architecture for: (a) $\beta$-VAE and (b) OF based OOD detectors.}
    \label{chained}
    \vspace{-7mm}
\end{figure}

Quantized solutions are deployed as ROS 2 callback graphs \cite{casini2019response} to the target hardware and evaluated against nonfunctional requirements. We consider the following ROS 2 callback graph architectures:
\begin{itemize}
    \item \textbf{Chained Multi-Threaded (Chain-MT)} - Each ROS node is launched as a separate Linux process. Fig.~\ref{chained} shows an example of the graphs for the OF and $\beta$-VAE detectors.  We break the detector into callbacks of roughly equal execution times for pipelined execution on the CPS.
    \item \textbf{Monolithic Single-Threaded (Mono-ST)} - A single OOD detection node is launched by the \textit{SingleThreadedExecutor}; all active callbacks share a single thread. 
    \item \textbf{Monolithic Multi-Threaded (Mono-MT)} - A single OOD detection node is launched by the \textit{MultiThreadedExecutor}; active callbacks are scheduled on a thread from a shared thread pool.
\end{itemize}

\section{Case Studies}
\subsection{Evaluation Setup}
\label{setup}

% Evaluation duckiebot platforms (DB-21M and modified DB-19) and environment.
We chose the Duckietown \cite{liam2017duckietown} platform to simulate AVs (i.e., DuckieBots) controlled by resource constrained hardware in a controlled environment (i.e., DuckieTowns). We selected two DuckieBots: the DB21M (Jetson Nano 2GB) and a modified DB19 (RPi 4B with Coral Edge TPU), and applied the \textit{PREEMPT\_RT} Linux kernel patch \cite{reghenzani2019real} to both. We ran experiments on the Jetson with GPU (Jetson-GPU), on the Jetson with only CPU (Jetson-CPU), on the RPi with TPU (RPi-TPU) and on the RPi with only CPU (RPi-CPU). We used ROS 2 to evaluate our OOD detectors since it is targeted at CPS and  provides the necessary infrastructure for scheduling callbacks and communication tasks.

% Insert table with the hardware specifications and environment
We gathered image data from a DuckieTown constructed out of an 8 meter long track with 5 different backgrounds in a laboratory environment (i.e., 5 scenes). For each scene, we collected images for four runs through the town.  We partitioned this data into in-distribution (ID) training and testing data.  We created OOD test images by digitally superimposing rain or snow and modifying ambient light conditions on non-training images. The script to perform this augmentation is available on GitHub\footnote{https://github.com/CPS-research-group/CPS-NTU-Public}.  We enforced a test set composition of 1/1 ID/OOD samples to avoid biasing our AUROC metric.

\subsection{Case Study: $\beta$-VAE OOD Detector}
\label{bvae}

\subsubsection{Gathering Requirements}
We attempted to detect OOD data arising from two generative factors: rain, and brightness.  ID training and calibration samples consisted of images with rain strength uniformly drawn between $0$ and $0.003$ and in-distribution brightness ranging from $-0.5$ to $0.5$.  We defined OOD samples for the rain partition as images with rain strength $0.004$ to $0.01$ or brightness outside the ID range.

\subsubsection{Model Development}
\begin{figure}[htbp]
    \centering
    \includegraphics[width=0.45\textwidth]{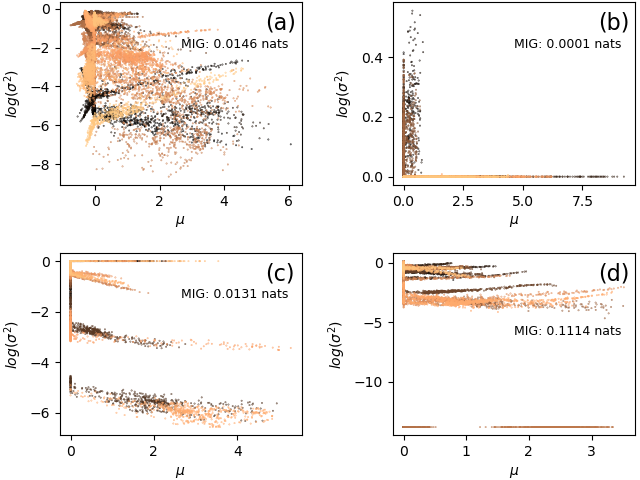}
    \caption{Visualization of ID latent distribution for several training configurations:  (a) Original model, leaky ReLU learning $\log(\sigma^2)$; (b) ReLU learning $\log(\sigma^2)$; (c) ReLU learning $-\log(\sigma^2)$; and (d) ReLU learning $\sigma^2$.}
    \label{fig:latent_dist}
    \vspace{-3mm}
\end{figure}

We implemented the $\beta$-VAE network as prescribed in \cite{ramakrishna2021efficient} with four convolutional layers of stride $1$, $3$x$3$ kernels, and depths 128/64/32/16.  Each convolutional layer was followed by a maxpooling layer with kernel $2$.  The convolutional layers were followed by $4$ fully connected layers with sizes 2048, 1000, 250, and $n_{latent}$ and the decoder was a mirror image of the encoder network.  A Bayesian optimization to maximize mutual information gain (MIG) \cite{ramakrishna2021efficient} yielded the hyperparameters $n_{latent}=36$ and $\beta=2.32$.  The calibration set for inductive conformal prediction (ICP) contained $3240$ ID images, to comply with the 2/1 training/calibration split recommended in \cite{ramakrishna2021efficient}.  Martingale window size was chosen as 20 since at a target data rate of 20 fps, the past second of predictions would be taken into consideration.  The optimal decay term in the cumulative summation was determined by a parameter sweep. The model proposed in~\cite{ramakrishna2021efficient} used leaky ReLU activation functions, however, the Coral Edge TPU did not support this function \cite{coraltpuops}, so we used ReLU instead.  Fig.~\ref{fig:latent_dist} (a) and (b) show the resulting latent distributions of the training set for the model with leaky ReLU and ReLU activation functions respectively.  (a) has achieved partial disentanglement, but (b) has not.  This is because in (a), most distributions have a negative log-variance, but ReLU cannot produce a negative value.  A disentangled latent space is required for the $\beta$-VAE OOD detector to function, so it is necessary to learn a parameter that can only take positive values.  Fig.~\ref{fig:latent_dist} shows the resulting latent space for ReLU learning $-\log(\sigma^2)$ (c) and $\sigma^2$ (d) along with their MIG scores.  We determined that configuration (d) had the highest MIG, even surpassing that of the original model.

\subsubsection{Satisfying Functional Requirements}
\label{bvae_tuning}
\begin{figure}[htbp]
    \centering
    \includegraphics[width=0.48\textwidth]{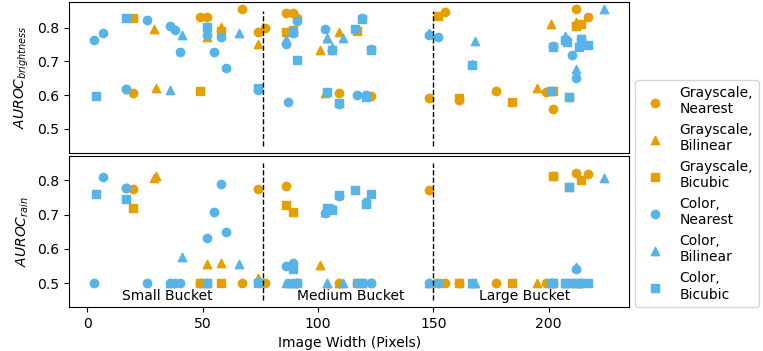}
    \caption{AUROCs for all solutions explored by GAs for $\beta$-VAE preprocessing.}
    \label{fig:bvae_ga}
    \vspace{-2mm}
\end{figure}

In the vanilla $\beta$-VAE OOD detector, RGB images are resized to $224$x$224$ with bilinear interpolation.  We setup three GAs whose fitness functions were the harmonic mean of AUROC for OOD brightness and rain samples.  The genes were: 1) \textit{image size}: the small bucket contained alleles for image widths 3px--76px, the medium bucket for 77px--150px, and the large bucket for 151px--224px; 2) \textit{interpolation}: nearest neighbor, bilinear, or bicubic; and 3) \textit{color space}: RGB or grayscale.  All three GAs ran for 16 generations at which point convergence occurred.  The AUROC values for all explored solution are shown in Fig.~\ref{fig:bvae_ga}.  It is interesting to note that solutions with a high AUROC for one data partition do not necessarily perform well at detection in another data partition. For example, when using 26x26 color images, the detector has an AUROC of 0.823 for brightness, but 0.5 for rain.  This provides justification for our use of harmonic mean across the different OOD partitions.  The variance in AUROC for images of similar sizes, or even the same size with different interpolation techniques is also interesting.  We observe that the average AUROC for the brightness partition was higher than that of rain, indicating that some generative factors may be more difficult to learn.

Table \ref{table:ga_bvae} shows the final fitness for the winners of each GA and the original solution (f32).  The winners of every GA were grayscale indicating that for these OOD factors, the extra color information was unnecessary for detection.  Furthermore, the medium and large buckets actually showed better performance with nearest neighbor interpolation, a simpler interpolation technique.  For larger image sizes, this may have had the effect of creating a sharper cutoff around rain drops, making detection easier.  The winner of the large bucket had a better fitness than the original solution.

We quantized the winning solutions to see what effect this would have on AUROC (Table \ref{table:ga_bvae}: f16, qint8, Coral).  For the smaller networks, qint8 improved the combined AUROC score.  One possible reason is the calibration set used in ICP had to be regenerated specifically for use with qint8.  Even though the immediate outputs of the encoder network had a limited number of values, the interaction with KL divergences from the calibration set in ICP and the martingale could have had an overall positive effect.  For half precision quantization, no re-calibration was performed, but the performance boost for smaller image sizes is still observed.  The quantization could be causing the existing network to generalize better.

\begin{table}[htbp]
    \centering
    \caption{Fitness of candidate solutions across different quantization schemes.} \label{table:ga_bvae}
    \begin{tabular}{|c|c|c|c|c|c|c|c|}  
    \hline
    \multirow{2}{*}{\textbf{GA}} & \multirow{2}{*}{\textbf{Size}} & \multirow{2}{*}{\textbf{Interp.}} & \multirow{2}{*}{\textbf{Color}} & \multicolumn{4}{c|}{\textbf{Fitness}} \\\cline{5-8}
    & & & & \textbf{f32} & \textbf{f16} & \textbf{qint8} & \textbf{Coral} \\
    \hline
    \textbf{S} & 29 & Bilinear & Gray &  0.802 & 0.812 & 0.817 & 0.817 \\
    \hline
    \textbf{M} & 86 & Nearest & Gray & 0.813 & 0.823 & 0.834 & 0.834 \\
    \hline
    \textbf{L} & 212 & Nearest & Gray & 0.837 & 0.837 & 0.836 & 0.836 \\
    \hline
    \textbf{Orig.} & 224 & Bilinear & RGB & 0.831 & 0.831 &  0.836 & 0.836 \\
    \hline
    
\end{tabular}
\vspace{-7mm}
\end{table}

\begin{figure*}[htbp]
    \centering
    \includegraphics[width=1\textwidth]{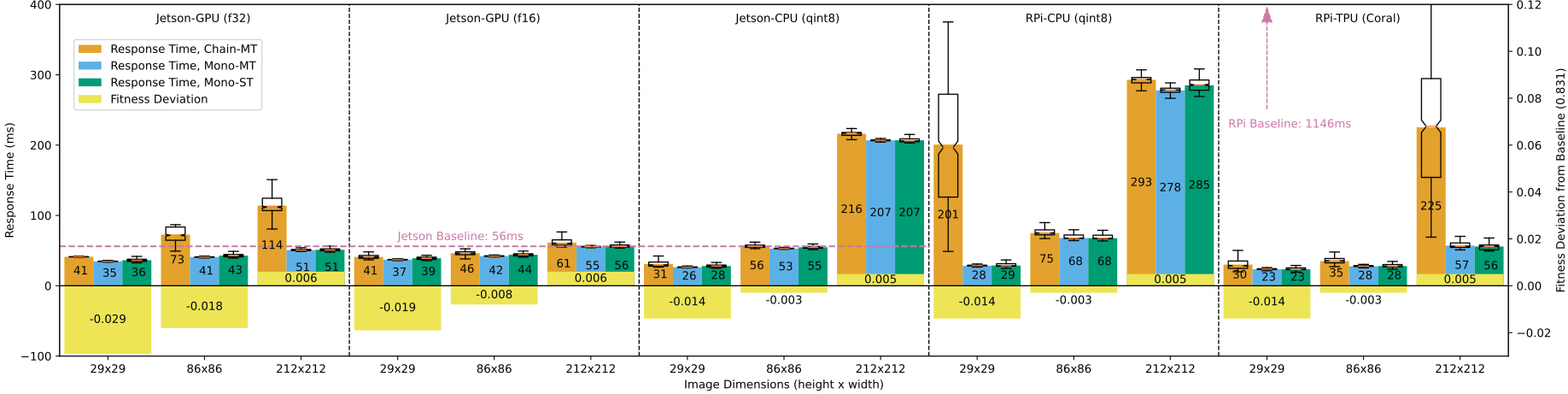}
    \caption{ Response times of the $\beta$-VAE OOD detector on DB21M (Jetson-GPU and Jetson-CPU) and DB19 (RPi-CPU and RPi-TPU) with different precisions (f32, f16, qint8, and Coral). The narrow bars show the mean response times for each callback graph, with box plots displaying the response time distributions. The fat bars show the deviation in AUROC from the baseline model.}
    \label{fig:bvae_times}
    \vspace{-5mm}
\end{figure*}

\begin{figure}[htbp]
\vspace{-3mm}
    \centering
    \includegraphics[width=0.4\textwidth]{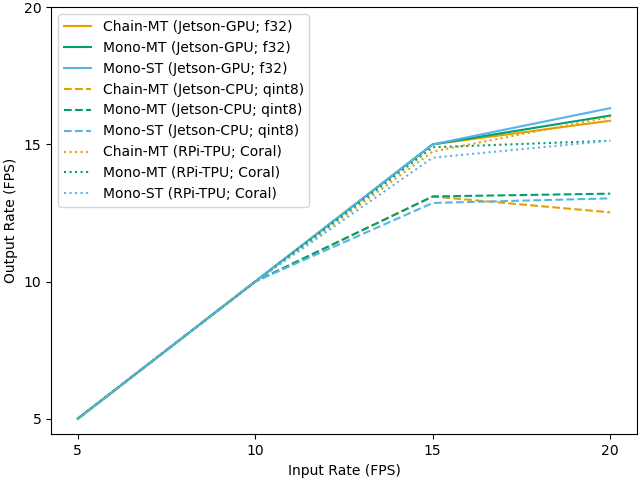}
    \caption{Sustained throughput of different ROS 2 callback graphs for the small (29x29) $\beta$-VAE OOD detector on Jetson-GPU, Jetson-CPU, and RPi-TPU.}
    \label{fig:bvae_tp}
\vspace{-6mm}
\end{figure}

\subsubsection{Satisfying Nonfunctional Requirements}
The quantized models for each candidate were then deployed to the Jetson and RPi for response time analysis as shown in Fig.~\ref{fig:bvae_times}.  The response times are compared to a baseline (f32 on GPU for Jetson and f32 on CPU for RPi) to show the decrease in response time from original model; the detectors with smaller inputs perform better in terms of response time.  When processing data on the GPU, the f16 model has a slower response time than f32, except for Chain-MT.  This is likely because a greater percentage of time is spent converting tensors to f16 in this network architecture.  In terms of response time, the qint8 model outperforms the GPU at smaller input sizes.  We observe that across all hardware and quantization types, the Chain-MT configuration has a slower response than Mono-ST or Mono-MT.  To see if the chained callback graph has the potential to increase throughput, Fig.~\ref{fig:bvae_tp} shows the throughput of the fastest OOD detectors from the Jetson-GPU (f32), Jetson-CPU (qint8), and RPi-TPU (Coral) configurations.  As the input fps rate to the detectors is increased, they eventually fail to sustain the same output rate and create a backlog.  The chained graph does not offer any advantage over the monolithic ones here; this may be due to all nodes spending time blocked while memory is transferred in and out of swap.  We believe the particularly large variance for Chain-MT on the RPi-TPU is exacerbated by the USB interface between the RPi and TPU.

\subsection{Case Study: Optical Flow OOD Detector}
\label{opt-flow}

\subsubsection{Gathering Requirements}
We used the first three runs through the DuckieTown for each scene (9,800 images) as training and validation data for the OF OOD detector, and withheld the final runs (3,190 images) for testing.  We digitally augmented half of the scenes in the test set with snow and rain at level 0.003 to maintain a 1/1 ID/OOD test split; any flows containing rain or snow were considered as OOD.

\subsubsection{Model Development}
The OF detector contains two encoder networks, both of which were implemented identically: four CNN blocks of 32/64/128/256 channels with 5x5 filters of stride $3$, ReLU activations, and BatchNorm  followed by one fully connected layer terminating with 12 latent variables. We trained and tested the encoders with PyTorch in f32. For f16 quantization, the models remained in PyTorch. For the qint8 and Coral evaluation, we converted the models via the model compression process in \cite{feng2021improving}.  The original model in \cite{feng2021improving} used the ELU activation function, however, the Coral TPU does not support this for quantization \cite{coraltpuops}, so the ReLU activation function was used instead. Unlike the $\beta$-VAE model, latent disentanglement is not a requirement.

\subsubsection{Satisfying Functional Requirements}
\begin{figure}[htbp]
\vspace{-4mm}
    \centering
    \includegraphics[width=0.41\textwidth]{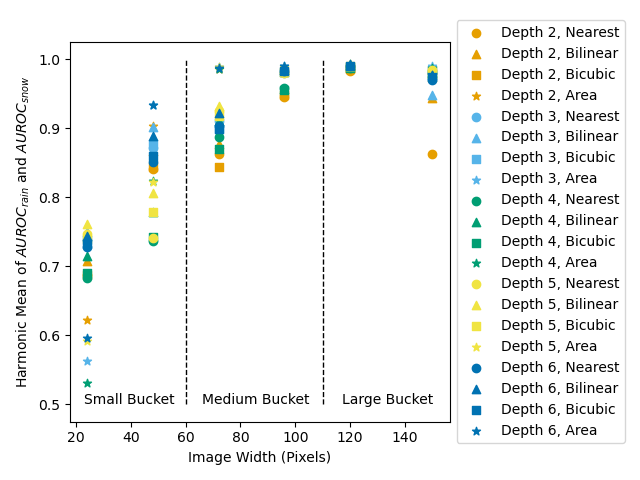}
    \caption{All solutions explored by the GA for optical flow.}
    \label{fig:of_ga}
    %\vspace{-3mm}
\end{figure}

In the original OF OOD detector, grayscale images were resized to $120$x$160$ with bilinear interpolation and sequential images were converted to horizontal and vertical flow matrices.  Subsequent flow matrices were concatenated to an input depth of 6.  We wanted to find a set of parameters that could reduce network inference time while maintaining comparable robustness. We used 3 GAs with following genes and alleles: 1) \textit{image size}: three buckets: small (24x32 and 48x64), medium (72x96 and 96x128) and large (120x160 and 150x200); 2) \textit{interpolation}: nearest neighbor, bilinear, bicubic, or pixel area relation; and 3) \textit{optical flow depth}: 2 to 6 sequential flow matrices stacked in the channel dimension of the input tensor. We selected the fitness function as the harmonic mean of AUROC for OOD rain and snow samples to find a detector that detects both factors well.  All three GAs ran for 100 generations to ensure convergence. Fig.~\ref{fig:of_ga} shows the solutions explored by the GA.  In general, OF depths of 6 with pixel area relation interpolation performed the best, except for input size 24x32. We select the best performing models (120x160x6, 96x128x6, and 48x64x6) from each bucket for quantization as shown in table~\ref{table:ga_of}.  In this detector, there is no AUROC boost from f16 and qint8 quantization as observed with $\beta$-VAE.

\begin{figure*}
\centering
    \includegraphics[width=1\textwidth]{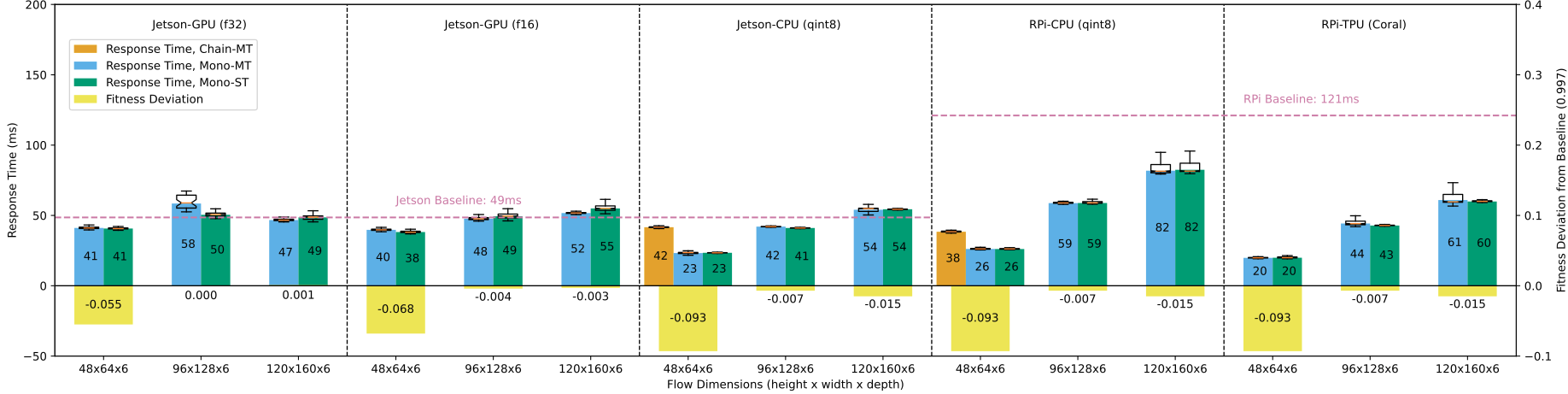}
    \caption{ Response times of the OF OOD detector on DB21M (Jetson-GPU and Jetson-CPU) and DB19 (RPi-CPU and RPi-TPU) with different precisions (f32, f16, qint8, and Coral). The narrow bars show the mean response times for each callback graph, with box plots displaying the response time distributions. The fat bars show the deviation in AUROC from the baseline model.}
    \label{fig:of_rt}
    \vspace{-6mm}
\end{figure*}

\begin{table}[htbp]
    \centering
    \caption{Fitness of candidate solutions across different quantization schemes. D. refers to optical flow depth.} \label{table:ga_of}
    \begin{tabular}{|c|c|c|c|c|c|c|c|}  
    \hline
    \multirow{2}{*}{\textbf{GA}} & \multirow{2}{*}{\textbf{Size}} & \multirow{2}{*}{\textbf{Interp.}} & \multirow{2}{*}{\textbf{D.}} & \multicolumn{4}{c|}{\textbf{Fitness}} \\\cline{5-8}
    & & & & \textbf{f32} & \textbf{f16} & \textbf{qint8} & \textbf{Coral} \\
    \hline
    \textbf{S} & 48x64 & Area & 6 & 0.942 & 0.929 & 0.904 & 0.904 \\
    \hline
    \textbf{M} & 96x128 & Area & 6 & 0.997 & 0.993 & 0.990 & 0.990 \\
    \hline
    \textbf{L} & 120x160 & Area & 6 & 0.998 & 0.994 & 0.982 & 0.982 \\
    \hline
    \textbf{Orig.} & 120x160 & Bilin. & 6 & 0.997 & 0.992 &  0.979 & 0.979 \\
    \hline
    
\end{tabular}
\vspace{-5mm}
\end{table}

\subsubsection{Satisfying Nonfunctional Requirements}
\begin{figure}[htbp]
\vspace{-2mm}
\centering
    \includegraphics[width=0.4\textwidth]{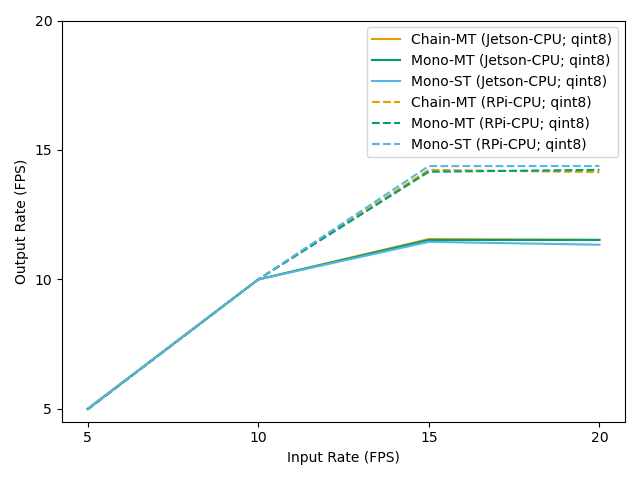}
    \caption{Sustained throughput of different ROS 2 callback graphs for the small (48x64x6) OF OOD detector on Jetson-CPU, and RPi-CPU. }
    \label{fig:of_throughput}
    \vspace{-5mm}
\end{figure}

Fig.~\ref{chained} (b) shows the computational steps of the OF OOD detector. Once an image arrives from the camera, the preprocessing steps (resize, sharpening, crop, and Farneback OF) must be run sequentially and cannot be parallelized. However, the output from the OF algorithm is fed to two encoder networks, and we can exploit parallelism here.  We construct the Chain-MT ROS 2 callback graph as shown in Fig.~\ref{chained} (b). In this scenario, we hope to exploit the available computational resources so that the two encoders run concurrently.  However, because the encoders for the f32, f16, and Coral models require a GPU/TPU and there is only one such available resource on either DuckieBot, we can only run the Chain-MT graph for the CPU (qint8) models as both systems have multi-core CPUs.  For the Mono-ST and Mono-MT graphs, all the quantization configurations were tested. The experimental results are shown in Fig.~\ref{fig:of_rt}.  We compare the top three models chosen by the GA search in phase 3 of our methodology with the baseline model (i.e., the 120x160x6/bilinear model) for their respective quantization levels. On the Jetson, there were no improvements for quantizing the 120x160x6 model and the Jetson-GPU (f32) model should be used as it gives the best AUROC. However, for the small and medium buckets across Jetson-GPU and Jetson-CPU, we achieved response time improvements varying from 2.0\% to 53.1\%, except for the f32-96x128x6 (-2.0\%), f16-120x160x6 (-6.1\%), f16-96x128x6 (on par), and qint8-120x160x6 (-10.2\%). In this case, the Jetson could comfortably run the f32 baseline model at 20 fps, and the qint8-48x64x6 model at 40 fps. However, this assumes that the Jetson is solely used for OOD detection and no other processes are competing with it for resources. On the RPi-CPU, running the baseline f32-120x160x6 model on Mono-ST for each inference took 121 ms on average, which translates to roughly 8 fps.  Compared to the baseline, we achieved response time improvements ranging from 32.2\% to 83.5\%.  In Fig.~\ref{fig:of_throughput}, we plotted the input fps rate to the detector versus its sustained output rate to see if Chain-MT could improve throughput.  Unfortunately, there was little difference between the ROS 2 callback graphs, although we note that the RPi outperformed the Jetson in terms of throughput.  We attribute this to Jetson's lack of memory, forcing callbacks to block while reading from and writing to swap/zram.

\section{Conclusion}
Designing an OOD detector for a CPS would have previously required evaluating many parameters to find a model that meets both functional and nonfunctional system requirements.  We proposed a methodology that simplifies the design process for any deep OOD detector.  It provides guidance on the selection of a preprocessing pipeline, quantization level, and ROS~2 callback graph for deployment to reduce response time over a baseline model while minimizing accuracy loss.  This was demonstrated on a $\beta$-VAE OOD detector where we obtained a 37.5\% response time reduction over the original model while maintaining 96.5\% of the original AUROC.  On the optical flow detector, our methodology obtained a 51.2\% response time reduction while maintaining 99.3\% of its original AUROC.  Future additions to this methodology could include analysis of memory usage and power consumption.  Furthermore, our experiments focused on scene-based OOD detection, but tests on object-based OOD detectors could yield additional insights.
%Previously designing an OOD detector for a CPS would have required evaluating many parameters to find a model that meets both the functional and nonfunctional system requirements.  We have proposed a framework that simplifies the design process on any DNN powered OOD detector.  It selects an preprocessing pipeline, quantization level, and ROS node graph deployment strategy to reduce response time over a baseline model while maintaining similar robustness.  This was demonstrated on a $\beta$-VAE OOD detector where we obtained a 40\% response time reduction over the original model while maintaining 97\% of the original AUROC.  On the optical flow detector, our framework obtained a 75\% response time reduction while maintaining 92\% of its original AUROC.  Future additions to this framework could include analysis of memory usage and power consumption.

\section*{Acknowledgment}
\vspace{-0.5mm}
This research was funded in part by MoE, Singapore, Tier-2 grant number MOE2019-T2-2-040.
\vspace{-1mm}
%\scriptsize
%\renewcommand*{\bibfont}{\footnotesize}
%\printbibliography


\begin{thebibliography}{00}
\bibitem{ramakrishna2021efficient}S. Ramakrishna, Z. Rahiminasab, G. Karsai, A. Easwaran, and A. Dubey, ``Efficient Out-of-Distribution Detection Using Latent Space of $\beta$-VAE for Cyber-Physical Systems,'' \textit{ACM Transactions on Cyber-Physical Systems (TCPS)}, vol. 6, no. 2, pp. 1--34, Apr. 2022, doi: 10.1145/3491243.
\bibitem{feng2021improving}Y. Feng, D. J. X. Ng, and A. Easwaran, ``Improving Variational Autoencoder based Out-of-Distribution Detection for Embedded Real-time Applications,'' \textit{ACM Transactions on Embedded Computing Systems (TECS)}, vol. 20, no. 5s, pp. 1--26, Oct. 2021, doi: 10.1145/3477026.
\bibitem{yuhas2021embedded}M. Yuhas, Y. Feng, D. J. X. Ng, Z. Rahiminasab, A. Easwaran, ``Embedded Out-of-Distribution Detection on an Autonomous Robot Platform,'' in \textit{Proceedings of the Workshop on Design Automation for CPS and IoT}, May 2021, pp. 13--18, doi: 10.1145/3445034.3460509.
\bibitem{hadidi2019characterizing}R. Hadidi, J. Cao, Y. Xie, B. Asgari, T. Krishna, and H. Kim, ``Characterizing the Deployment of Deep Neural Networks on Commercial Edge Devices,'' in \textit{2019 IEEE International Symposium on Workload Characterization (IISWC)}, Nov. 2019, pp. 35--48, doi: 10.1109/IISWC47752.2019.9041955.
\bibitem{zhang2019openei}X. Zhang, Y. Wang, S. Lu, L. Liu, L. Xu, and W. Shi, ``OpenEI: An Open Framework for Edge Intelligence,'' in \textit{2019 IEEE 39th International Conference on Distributed Computing Systems (ICDCS)}, Jul. 2019, pp. 1840--1851, doi: 10.1109/ICDCS.2019.00182.
\bibitem{stacker2021deployment}L. Stäcker \textit{et al.}, ``Deployment of Deep Neural Networks for Object Detection on Edge AI Devices with Runtime Optimization,'' in \textit{2021 IEEE/CVF International Conference on Computer Vision Workshops (ICCVW)}, Oct. 2021, pp. 1015--1022, doi: 10.1109/ICCVW54120.2021.00118.
\bibitem{thar2019meta}K. Thar, T. Z. Oo, Z. Han, and C. S. Hong, ``Meta-Learning-Based Deep Learning Model Deployment Scheme for Edge Caching,'' in \textit{2019 15th International Conference on Network and Service Management (CNSM)}, Oct. 2019, pp. 1--6, doi: 10.23919/CNSM46954.2019.9012733.
\bibitem{berthelier2021deep}A. Berthelier, T. Chateau, S. Duffener, C. Garcia, and C. Blanc, ``Deep Model Compression and Architecture Optimization for Embedded Systems: A Survey,'' \textit{Journal of Signal Processing Systems}, vol. 93, no. 8, pp. 863--878, Aug. 2021, doi: 10.1007/s11265-020-01596-1.
\bibitem{hendrycks2016baseline}D. Hendrycks and K. Gimpel, ``A baseline for detecting misclassified and out-of-distribution examples in neural networks,'' 2016, \textit{arXiv:1610.02136}.
\bibitem{cass2019taking}S. Cass, ``Taking AI to the edge: Google's TPU now comes in a maker-friendly package,'' \textit{IEEE Spectrum}, vol. 56, no. 5, pp. 16--17, May 2019.
\bibitem{coraltpuops}``TensorFlow models on the Edge TPU,'' Coral.ai.  https://coral.ai/docs/edgetpu/models-intro/ (accessed Jun. 23, 2022).
\bibitem{casini2019response}D. Casini, T. Bla{\ss}, I. L{\"u}tkebohle, and B. Brandenburg, ``Response-Time Analysis of ROS 2 Processing Chains Under Reservation-Based Scheduling,'' in \textit{31st Euromicro Conference on Real-Time Systems (ECRTS 2019)}, Jul. 2019, pp. 1--23, doi: 10.4230/LIPIcs.ECRTS.2019.6.
\bibitem{liam2017duckietown}L. Paull \textit{et al.}, ``Duckietown: An open, inexpensive and flexible platform for autonomy education and research,'' in \textit{2017 IEEE International Conference on Robotics and Automation (ICRA)}, May 2017, pp. 1497--1504, doi: 10.1109/ICRA.2017.7989179.
\bibitem{reghenzani2019real}F. Reghenzani, G. Massari, and W. Fornaciari, ``The Real-Time Linux Kernel: A Survey on PREEMPT\_RT,'' \textit{ACM Computing Surveys (CSUR)}, vol. 52, no. 1, pp. 1--36, Jan. 2020, doi: 10.1145/3297714.
\end{thebibliography}
\end{document}